\documentclass{article}

\usepackage{arxiv}

\usepackage[pdftex]{graphicx}
\usepackage{url}
\usepackage{subfig}
\usepackage{verbatim}
\usepackage{nth}
\usepackage{soul}
\usepackage{amsmath}

\usepackage[utf8]{inputenc} % allow utf-8 input
\usepackage[T1]{fontenc}    % use 8-bit T1 fonts
\usepackage{hyperref}       % hyperlinks
\usepackage{booktabs}       % professional-quality tables
\usepackage{amsfonts}       % blackboard math symbols
\usepackage{nicefrac}       % compact symbols for 1/2, etc.

\AtBeginDocument{%
  \providecommand\BibTeX{{%
    \normalfont B\kern-0.5em{\scshape i\kern-0.25em b}\kern-0.8em\TeX}}}

\begin{document}

\title{Controllable Exploration of a Design Space\\ via Interactive Quality Diversity}

\author{
    \href{https://orcid.org/0000-0002-4363-4922}{\includegraphics[scale=0.06]{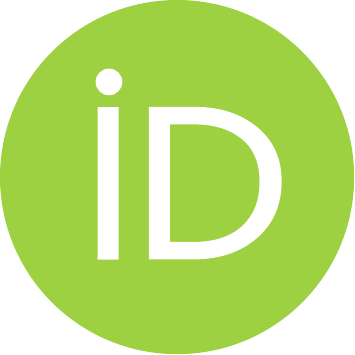}\hspace{1mm}Konstantinos Sfikas} \\
	Institute of Digital Games\\
	University of Malta\\
	Msida, Malta\\
	\texttt{konstantinos.sfikas@um.edu.mt} \\
	%% examples of more authors
	\And
	\href{https://orcid.org/0000-0001-5554-1961}{\includegraphics[scale=0.06]{orcid.pdf}\hspace{1mm}Antonios Liapis} \\
	Institute of Digital Games\\
	University of Malta\\
	Msida, Malta\\
	\texttt{antonios.liapis@um.edu.mt} \\
	\And
	\href{https://orcid.org/0000-0003-2728-4026}{\includegraphics[scale=0.06]{orcid.pdf}\hspace{1mm}Georgios N. Yannakakis} \\
	Institute of Digital Games\\
	University of Malta\\
	Msida, Malta\\
	\texttt{georgios.yannakakis@um.edu.mt} \\
}

\renewcommand{\shorttitle}{Controllable Exploration of a Design Space via Interactive Quality Diversity}

\date{}

\maketitle

\begin{abstract}

This paper introduces a user-driven evolutionary algorithm based on Quality Diversity (QD) search. During a design session, the user iteratively selects among presented alternatives and their selections affect the upcoming results. We aim to address two major concerns of interactive evolution: (a) the user must be presented with few alternatives, to reduce cognitive load; (b) presented alternatives should be diverse but similar to the previous user selection, to reduce user fatigue. To address these concerns, we implement a variation of the MAP-Elites algorithm where the presented alternatives are sampled from a small region (window) of the behavioral space. After a user selection, the window is centered on the selected individual’s behavior characterization, evolution selects parents from within this window to produce offspring, and new alternatives are sampled. Essentially we define an adaptive system of local QD, where the user’s selections guide the search towards specific regions of the behavioral space. The system is tested on the generation of architectural layouts, a constrained optimization task, leveraging QD through a two-archive approach. Results show that while global exploration is not as pronounced as in MAP-Elites, the system finds more appropriate solutions to the user's taste, based on experiments with controllable artificial users.
\end{abstract}


\section{Introduction}\label{sec:introduction}

Interactive Evolutionary Computation (IEC) is a form of ``systemic optimization that uses a real human’s subjective evaluation in its optimization process'' \cite{pei2018researchPS}. In its narrow definition, the user's subjective evaluation essentially takes the role of fitness in an Evolutionary Computation (EC) optimization process \cite{takagi2001interactive}. The main benefit of IEC is that it is applicable to problems where the definition of a fitness function is hard, or even impossible, while their evaluation from real humans is feasible. IEC however can easily lead to user fatigue \cite{takagi2001interactive}: EC typically operates better with large populations and many generations, while a human's ability to examine and compare solutions is a valuable and limited resource. Several solutions have been proposed to address user fatigue; among the most promising are function approximation via user models \cite{liapis2012adaptivemodel,hagg2019modeling}, which allows evolution to run for many generations in-between user interactions, and interactive differential evolution \cite{pei2018researchPS}. However, training a user model as an approximator might be slow to adapt to the user's current interactions \cite{liapis2013rie}, while population-based metaheuristics are suited to real-valued continuous genotypic representations which may not always be possible.

In this paper we introduce a novel IEC algorithm aiming to provide a high degree of user control, with a reduced degree of user fatigue. We showcase that this is achievable by exploiting the illumination capabilities of Quality Diversity (QD) algorithms \cite{pugh2016qualityDA}. We envision a hybrid system of ``User-Controlled QD Exploration'', where the user's subjective criteria localize and control QD search within a part of the behavioral space. We implement this concept by modifying the basic operation of MAP-Elites \cite{mouret2015illuminatingSS}, a popular QD algorithm, and devising the following interaction loop: we constrain the algorithm's operation within a window that covers a small region of the feature map, where it locally expands the archive for a number of generations. Afterwards, design alternatives are sampled from within the window and presented to the designer. Finally, the user's selection determines where the window will move towards next.
These steps summarize the functionality of the specific algorithmic implementation introduced here, which we refer to as User Controlled MAP-Elites (UC-ME).

We test UC-ME on the constrained design problem of generating architectural layouts, using a generative methodology that was introduced in \cite{sfikas2022aGP}. This problem has multiple constraints, an ad-hoc representation and ad-hoc genetic operators. To apply UC-ME on a constrained problem we draw inspiration from FI-MAP-Elites \cite{sfikas2022aGP} and adapt UC-ME to work on the dual archives of feasible and infeasible elites. To test how UC-ME caters to different potential user goals, we utilize artificial users with different selection criteria.

\section{Related Work}\label{sec:relatedwork}

The following sections highlight research on topics directly relevant to our work: Interactive Evolution and Quality Diversity.

\subsection{Interactive Evolutionary Computation}

As discussed in Section \ref{sec:introduction}, IEC is a powerful tool for domains where formulating a fitness function is as difficult as the problem itself. IEC has been applied for the generation of abstract images \cite{secretan2011picbreederAC}, 3D models \cite{clune2013uploadAO}, and sounds \cite{hoover2015interactivelyEC}, and its process has also been gamified, i.e. fully implemented within games as a part of the games' mechanics \cite{risi2016petalzSB, hastings2009galacticAR, soule2017darwinsDD}.

Several proposals have been put forth to address user fatigue in IEC \cite{takagi2001interactive}, e.g. by showing a subset of the population to the user, evolving through surrogate fitness functions \cite{liapis2012adaptivemodel} for more than one generation, and using online tools to share the burden of IEC with multiple users \cite{hastings2009galacticAR,secretan2011picbreederAC,risi2016petalzSB}. As a specific example of surrogate fitnesses for IEC, past work has proposed computational models of the designer \cite{liapis2013designermodeling, liapis2014modelingsketchbook} to drive evolution. Relevant work in \cite{liapis2014modelingsketchbook} showcases that designer models that are updated after every human interaction with the tool can be used as a fitness function in intermediate generations (between user selections), thus reducing user involvement. The main drawback of this approach is that it sacrifices the immediacy of user control, and must balance between overfitting or underfitting to the user's latest expressed preferences.

Other methods attempt to address the issue of user fatigue by accelerating the evolutionary process and its convergence to an optimum. Those include fitness landscape approximation \cite{pei2011acceleratingEC, pei2012fourierAO}, interactive differential evolution \cite{takagi2009pairedCB, pei2013tripleAQ, pei2017localFL} and convergence point estimation \cite{murata2015analyticalEO}, among others. The main limitation of these methods is that they are mostly applicable to problems whose genotypic representation is (or can be) expressed as a numeric vector which defines a continuous space. However, design problems (such as the use case examined in this paper) often require the use of ad-hoc genotypic representations that would be hard, if not impossible, to be reduced to a vector of real numbers.

\subsection{Quality Diversity}

Quality Diversity (QD) search \cite{pugh2016qualityDA} is a form of EC that simultaneously optimizes and diversifies the population of generated solutions. Exemplary algorithms of this approach include Novelty Search with Local Competition \cite{lehman2011evolvingAD} and MAP-Elites \cite{mouret2015illuminatingSS}. 
The original aim of QD algorithms was to tackle deception in EC, by ensuring population diversity along dimensions detached to the problem's fitness.
Their ability to illuminate a problem space, however, has been recognized as a way to foster human creativity \cite{gravina2019pcgqd, liapis2013transformingEC}.
QD search has thus spread beyond its original application in evolutionary robotics \cite{mouret2015illuminatingSS, lehman2011evolvingAD} to various design-related problems  \cite{khalifa2018talakatBH, 
alvarez2019empowering, 
galanos2021archelites}.

MAP-Elites \cite{mouret2015illuminatingSS} is one of the most well-researched QD algorithms. The algorithm operates by subdividing a multidimensional feature space into cells. Each cell may contain a single individual, thus forming an archive of elites within their own behavioral niches. During its operation, MAP-Elites randomly selects individuals from the occupied cells, mutates them and generates offspring. The offspring is assigned an appropriate behavioral niche, defined by multiple Behavioral Characterizations (BCs) acting as coordinates of the feature map. The offspring is placed in the cell for that niche, if that cell is currently empty; if the cell is already occupied, the offspring replaces the elite in that cell if its fitness is higher. Over many repetitions of this process, the algorithm both expands the archive (towards all possible BCs) and improves the fitness of elites contained therein.

\subsubsection{Constrained Quality Diversity}
When a problem includes hard constraints, solutions are characterized as feasible or infeasible if they satisfy a set of criteria or not. Importantly, their feasibility often requires special treatment, as it is often unrelated to fitness and may require the simultaneous satisfaction of different constraints. Although these types of problems are not the main concern of QD algorithms, various approaches exist in the literature \cite{liapis2015constrainedNS, khalifa2018talakatBH} that combine QD with constraint solving. Many of these methods draw inspiration from the FI-2-Pop GA \cite{kimbrough2008onAF} for their constraint solving aspects. In short, the FI-2-Pop GA operates on two populations, one containing feasible solutions and one containing infeasible. Parents are selected by either population, cross-breed only with other parents from the same population, but may produce offspring that are assigned to the other population, depending on their feasibility.

In constrained QD search, FI-MAP-Elites \cite{sfikas2022aGP} hybridizes FI-2Pop GA \cite{kimbrough2008onAF} and MAP-Elites \cite{mouret2015illuminatingSS} by maintaining two archives (one with feasible elites and one with infeasible elites). Parent selection alternates between the two archives, while mutated offspring can change archives based on their feasibility. Quality in the infeasible archive is based on how close the individual is to being feasible. BCs do not need to be shared between the two archives (although they often are). More details on the algorithm can be found in \cite{sfikas2022aGP}.

\subsubsection{Interactive Quality Diversity}

Despite its high suitability for design problems, QD in a Mixed-Initiative setting \cite{liapis2016mixedinitiative} is relatively under-researched. Notable examples can be found in the works of Alvarez et al. on Interactive, Constrained Quality Diversity \cite{alvarez2019empowering, alvarez2020interactive}. In these works, the user's interaction is implemented in two ways: (a) the user can control the MAP-Elites algorithm's parameters in a dynamic way, thus allowing them to illuminate the design space in different ways. Second, at a low-level, the designer can manually intervene on the generated designs, or design their own from scratch and then use them as seeds for MAP-Elites, to automatically generate variations. It is important to emphasize that while these modes of operation rely on a user's initiative, they are not matching the narrow definition \cite{pei2018researchPS} of IEC (see Section \ref{sec:introduction}).

\section{User-Controllable MAP-Elites} \label{sec:interactive_fi_map_elites}

We introduce the User Controllable MAP-Elites (UC-ME) algorithm as a way to endow a user with control over the direction and computational resources of the QD exploration. The algorithm operates by allowing parent selection only from a small window of the archive of elites, and moving this \textit{selection window} according to the user's selections. The user can select one favorite solution among a small number of alternatives (four in this paper) that have been sampled from within the current window. The general process of the UC-ME algorithm, and different methods for sampling design alternatives to show the user, are described in this paper.

\subsection{Algorithm Initialization} \label{sec:algorithm_initialization} 

UC-ME starts by producing a number of initial individuals through a random initialization method and placing them in the MAP-Elites archive according to their behavioral characterization. This step seeds the archive to enable interaction with the human user. The initial selection window of size $w{\times}w$ is centered at the cell with the mean BC values of existing elites, or the nearest elite if that cell is unoccupied.
The window size ($w$) is a parameter of UC-ME which should be much smaller than the resolution of the feature map.

\subsection{Interactive Operation} \label{sec:interactive_operation} 

After the algorithm has been initialized, the interactive session can begin. During an interactive session, the following steps are repeated indefinitely, until the designer decides to end it. 
\begin{enumerate}
    \item \textbf{Design Alternatives Sampling:}
    The algorithm samples $D$ design alternatives, from within the selection window, to be shown to the designer as options to select from.
    \item \textbf{Designer Input:}
    The designer selects one preferred design.
    \item \textbf{Selection Window Placement:}
    The selection window is centered at the coordinates of the designer's last selection.
    \item \textbf{Windowed Archive Expansion:}
    The algorithm operates for $N_e$ evaluations, selecting parents from within the window.
    The mutated offspring are evaluated and placed at their corresponding archive cell, based on their Behavioral Characterization coordinates, without being constrained by the window. 
    In case an offspring lands on an already occupied cell, the individual with the highest fitness survives.
\end{enumerate}

\subsection{Design Alternatives Sampling Methods} \label{sec:interactive_design_alternatives}

\begin{figure}
\centering
\includegraphics[width=.58\textwidth]{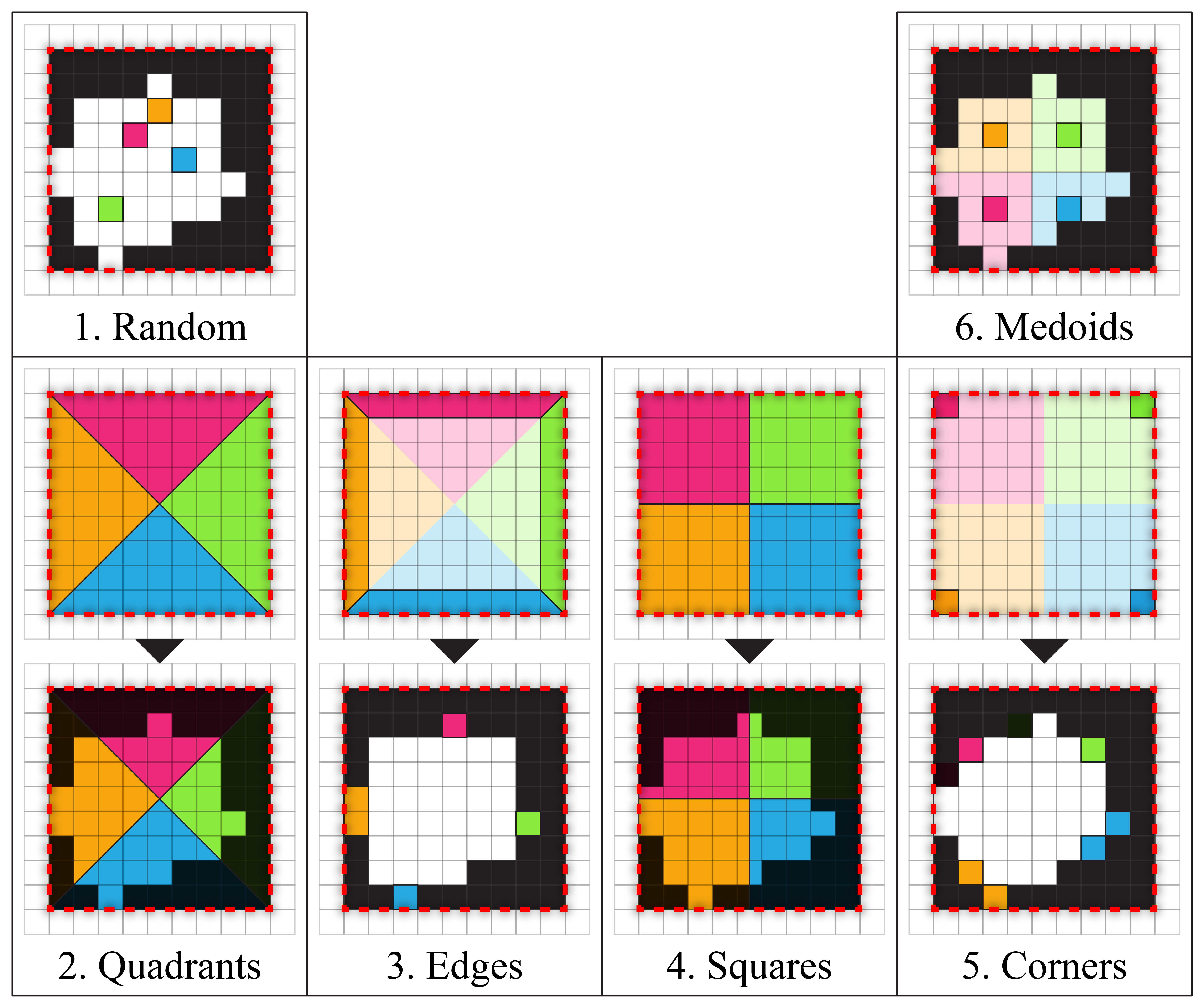}
\caption{Visual representation of the available DAS methods, applied to a selection window of $9 \times 9$ cells, for a hypothetical underlying coverage (in white). In case there are multiple cells of the same color, one candidate is chosen randomly from cells of the same color.}
\label{fig:interactive_design_alternatives}
\end{figure}

As described in Section \ref{sec:interactive_operation}, the algorithm samples a number of design alternatives to present to the user, from within the selection window. We only test UC-ME with four alternatives in this paper, and examine six methods for design alternatives sampling (DAS), which are all stochastic to some degree.
The following list describes each DAS method; Fig. \ref{fig:interactive_design_alternatives} visualizes a hypothetical DAS example.
\begin{itemize}
    \item \textbf{Random ($A_R$)} samples 4 elites from the window at random.
    \item \textbf{Quadrants ($A_Q$)} and \textbf{Squares ($A_S$)} split the window into 4 equal sections, using the diagonals ($A_Q$) or the $x$-$y$ axes ($A_S$). One individual is sampled randomly from each section.
    \item \textbf{Edges ($A_E$)} samples one individual at random from each of the 4 edges of the window. If no individual is on the edge, then the nearest individuals to that edge are preferred.
    \item \textbf{Corners ($A_C$)} samples one individual per corner of the window, or the nearest individual to that corner (as in $A_E$).
    \item \textbf{Medoids ($A_M$)}:
    The coordinates of the individuals within the selection window are used as data points in a $k$-medoids clustering algorithm, where $k=4$ in this paper. The four medoids of these clusters are shown to the user.
\end{itemize}

\section{Use Case: Layout Generation}\label{sec:usecase}

We test UC-ME on the complex and constrained problem of generating architectural layouts. Variations of this task can be found in computational architecture \cite{Charman1994ConstraintBasedAF, Medjdoub2000SeparatingTopologyAG, Rodrigues2013EvolutionaryStrategyEWp1} and in games \cite{karavolos2016evolving, dormans2010adventures}.
%,smith2018graphbased}.
We follow the methodology of \cite{sfikas2022aGP}, where the problem definition is a set of topological and other constraints, and the output is a geometrical solution that respects these constraints. We summarize the process for this use case below; more details can be found in \cite{sfikas2022aGP}.

We chose to focus on this complex problem for two reasons: first, architectural layouts are characterized by many quantifiable, yet subjective, features, making them ideal for testing MI-CC methods. Second, this task offers an opportunity to test the proposed methodology on a constrained domain, showcasing its extensibility to other MAP-Elites variants.

\subsection{Layout Representation}\label{sec:state_representation}

The representation of the architectural layout has two facets: a Design Specification (DS) and a Design Implementation (DI). 

The DS is a user-defined description of the problem at hand. It contains a connectivity graph of the layout, where vertices represent space units (rooms or other regions) and edges represents a connection between two space units through a door (see e.g. Fig. \ref{fig__connectivity_graph}). It also includes the desired area per space-unit, the number of doors to the exterior and windows in each space-unit, and whether it is an indoors or outdoors space (see e.g. Table \ref{tab__space_units}). 

The DI is a concrete, geometric implementation of a DS, where every space unit occupies a specific region of the plane, having a specific shape. The DS also includes the precise location of doors and windows. In order to support the generation of free-form geometries (i.e. not confined to fixed geometric systems such as grids), we implement a system based on a Voronoi-tessellation of the plane. The generated space units can be placed at specific regions of this tessellation. Both the space-units' placement and the underlying structure are mutated during the algorithm's operation.

\subsection{Constraints}\label{sec:constraints}

Architectural layouts are highly constrained problems, both from the physical requirements and design constraints (from the DS). We consider layouts as \emph{feasible} if (a) each space units take up a single, connected region; (b) adjacent space units in the DS share a boundary; (c) the area of each space unit is suitably close to its specified area in the DS; (d) all prescribed doors and windows are properly placed and (e) all interior pathways are at least 0.5 meters wide; (f) the entire Voronoi graph is connected and (g) at least 50\% of the Voronoi cells do not touching the border. The latter two constraints are low-level engine constraints to ensure that the generation process for space units is possible. Every generated individual is tested against these constraints, and if it fails any of them it is assigned a \emph{feasibility score} proportionate to the number of constraints succeeded, and the degree that they are succeeded.

\subsection{Fitness}\label{sec:fitness}

In this architectural layout problem, quality is mainly ensured by the satisfaction of constraints. Any individual that satisfies the constraints of Section \ref{sec:constraints} is ultimately ``good enough'' as a layout. In order to guide the QD algorithm, however, we use the adherence to the DS as our main quality criterion in the feasible population and optimize how close the areas of the space units are to the specified ones. We define mean area precision ($\bar{P}_s$) as the average difference between specified and actual area of each space unit. For each space unit its area precision is calculated via Eq.~\eqref{eq:area_precision}; If a space unit from the DS is missing from the DI then its $P_s$ is 0. Note that $\bar{P}_s$ is also a criterion for feasibility: indeed, if $\bar{P}_s<0.6$ then the individual is infeasible. If $\bar{P}_s\geq0.6$, this metric is treated as quality characterization for the feasible archive.
\begin{equation}
P_s=
\begin{cases}
A / A_t & A < A_t \\
A_t / A & A_t \leq A \\
\end{cases}\label{eq:area_precision}
\end{equation}
\noindent where $A$ is the area of the generated space unit and $A_t$ is the target area for this space unit prescribed in the DS.

\subsection{Behavioral Characterizations}\label{sec:behavioral_characterizations}
We use two Behavioral Characterizations (BCs) as measures of diversity of generated DIs. These BCs are used in both the feasible and infeasible archives and their definition follows:

{\textbf{Mean Space Units' Compactness}} ($\bar{C}_s$) is based on the notion of compactness, a unit-less measure that expresses the relation between a shape's perimeter and its area \cite{koutsolampros2019dissectingVG}.  A single space unit's compactness ($C_s$) is calculated as  via Eq. \eqref{eq:space_unit_compactness}; If a space unit is defined in the DI but is missing from the DS, its $C_s=0$. Finally, $\bar{C}_s$ measures the mean compactness of all space units in the DI.
\begin{equation}\label{eq:space_unit_compactness}
C_s = 2 \pi A / \Pi ^ 2
\end{equation}
\noindent where $A$ is the space unit's surface area and $\Pi$ its perimeter.

{\textbf{Plan Orthogonality}} ($\bar{O}_\theta$) is calculated as the mean orthogonality of all angles ($\theta$) between connected walls in the layout. A single angle's orthogonality ($O_\theta$) is calculated as shown in Eq. \eqref{eq:angle_orthogonality};
For acute angles, $O_\theta$ increases linearly from $0$ to $1$. For obtuse angles $\theta \in [\pi/2, 3\pi/4]$, $O_\theta$ decreases linearly from $1$ to $0.5$. For obtuse angles $\theta \in [3\pi/4, \pi]$, $O_\theta$ increases linearly from $0.5$ to $1$.
\begin{equation}\label{eq:angle_orthogonality}
O_\theta=
\begin{cases}
2\theta / \pi                & 0 \leq \theta < \pi/2 \\
2 - 2\theta / \pi            & \pi/2 \leq \theta < 3 \pi / 4 \\
2\theta / \pi -1           & 3 \pi / 4 \leq \theta \leq \pi \\
\end{cases}
\end{equation}
\noindent where $\theta$ represents an unsigned angle between two continuous line segments, always in the range of $[0\dots\pi]$.

\subsection{Initial Generation} \label{sec:initial_generation}

The initial generation of layouts occurs in a semi-stochastic manner that does not guarantee feasibility. 

First, a set of points are randomly placed within the bounding rectangle, and define a Voronoi tessellation of the plane; then, the prescribed space-units in the DS are iteratively placed on the plane starting from space units with the most connections; each space unit is assigned a number of cells that matches the area prescribed in the DS. Then, if possible, prescribed openings are placed randomly on walls in each room.

\subsection{Genetic Operators}\label{sec:operators}
Mutation of layouts occurs in two stages, stochastic \emph{destruction} and scripted \emph{repair}. This approach produces offspring that have partial similarities with the parents (as only parts of the layout is destroyed), while the repair functions aims to decrease the chance of producing infeasible offspring.

In the destruction step, the algorithm selects and applies between 1 and 3 of the following stochastic operations: (1) moving all points of the underlying Voronoi grid along the same direction; (2) moving a small set of Voronoi points in random directions; (3) deleting an entire room; (4) expanding an existing room, either taking over cells from other rooms (unsafe expansion) or respecting them (safe expansion); (6) eroding the exteriors cells of a room while still maintaining connectivity; (7) deleting a subset of existing openings.

During the repair step, the following operations are applied sequentially: (1) adding any missing rooms, if there is enough space; (2) re-connecting segments of the same room that may have become disconnected; (3) expanding rooms in order to fulfill adjacency requirements with other rooms; (4) deleting or adding room segments until they are suitably close to the desired area in the DS; (5) deleting extraneous openings and placing missing openings.  

\subsection{Constrained QD Process}

Due to the constrained nature of this problem, we modify the general operation of UC-ME (Section \ref{sec:interactive_fi_map_elites}) inspired by the constrained FI-MAP-Elites \cite{sfikas2022aGP} algorithm.

Constrained UC-ME works on a two-archives approach: one archive for the feasible elites and one for infeasible. We employ two BCs, described in Section \ref{sec:behavioral_characterizations}, and apply them to both archives. For the infeasible archive, we assign elites a quality based on their feasibility score (see Section \ref{sec:constraints}), while for the feasible archive we assign quality as $\bar{P}_s$. We modify the initialization process of Section \ref{sec:algorithm_initialization} as follows: the 100 initial individuals are assigned to the feasible or infeasible archive, according to their feasibility. Afterwards, we run FI-MAP-Elites QD search until the feasible archive has at least 1\% coverage. This provides us with enough of a seed to run the interactive operation of the algorithm; in terms of this operation, the only changes are that (a) DAS methods are applied only on the feasible archive, (b) the selection window is applied to both the feasible and the infeasible archive and (c) parents are chosen in an alternating fashion between the feasible and the infeasible archive. As in the basic UC-ME of Section \ref{sec:interactive_operation}, offspring are tested for feasibility and their BCs and placed in the appropriate archive in the appropriate cell, replacing any elite there if they have a higher feasibility (for the infeasible archive) or a higher $\bar{P}_s$ score (for the feasible archive).

\section{Experiment Protocol}\label{sec:protocol}

As a specific case study for architectural layout generation, we set up an experiment with a specific design specification for a medium-size apartment (Section \ref{sec:protocol_case}), algorithm parameters (Section \ref{sec:protocol_parameters}), controllable (artificial) users to test the algorithm (Section \ref{sec:protocol_users}). We also identified a plethora of performance metrics in order to assess the general and user-specific efficacy of the algorithm, which are described in Section \ref{sec:protocol_metrics}.

\subsection{Apartment Layout Specification}\label{sec:protocol_case}
\begin{figure}
\centering
\includegraphics[width=.4\textwidth]{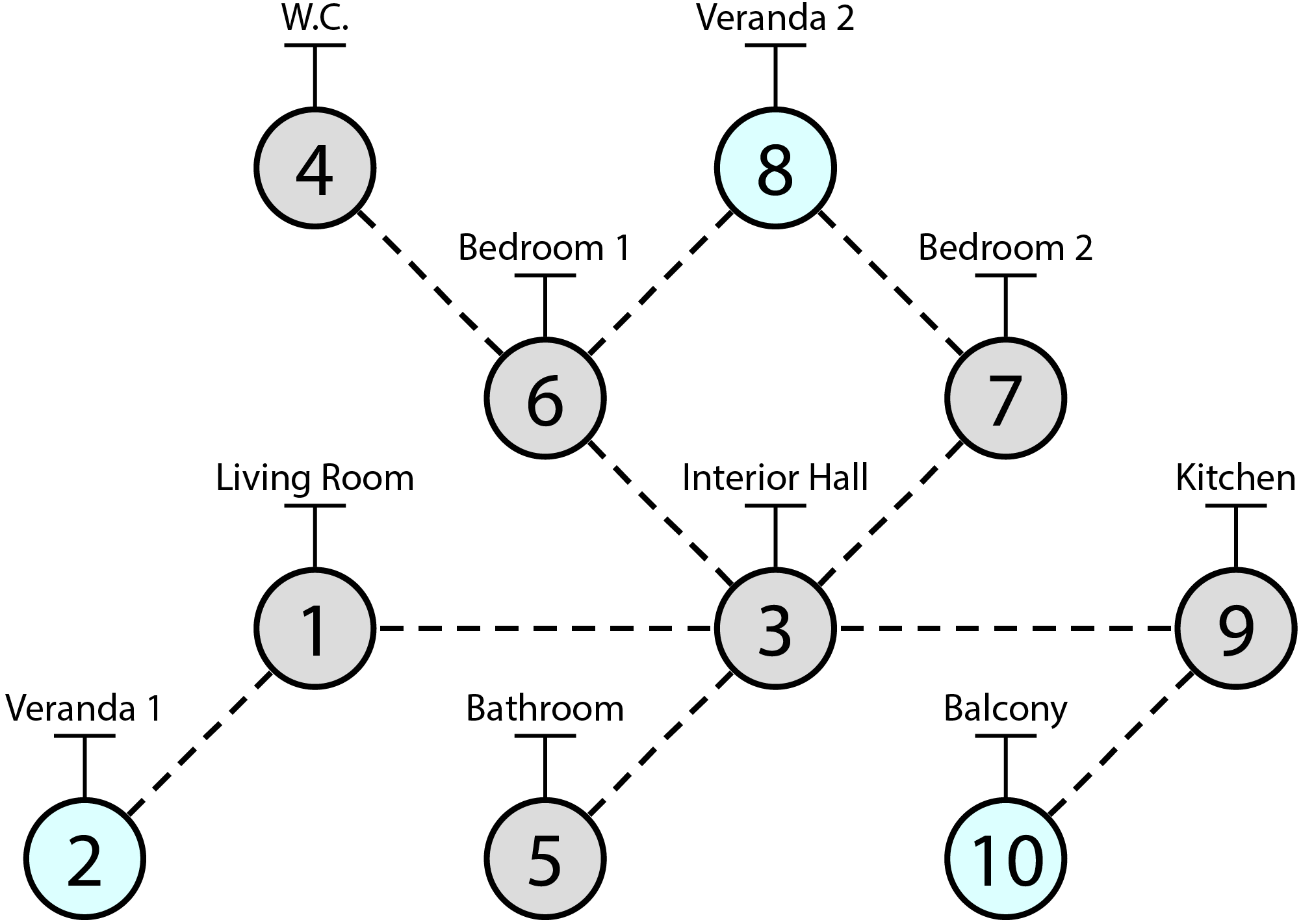}
\caption{Connections between space-units in the DS. Interior Space Units are in gray and exterior ones in cyan.}
\label{fig__connectivity_graph}
\end{figure}

For this paper we use a DS of a typical Mediterranean apartment. The DS consists of 10 space units, as described in Table \ref{tab__space_units}. The prescribed connections between them are visualized in Fig. \ref{fig__connectivity_graph}.

\begin{table}
\small
    \centering
    \begin{tabular}{l|l|c|c|c|c}
        % \hline
        % \multicolumn{5}{|c|}{Space Units and their properties} \\
        %\hline
        ID & Name          & Type     & Area     & Entrances & Windows \\ 
        \hline
        1  & Living Room   & Interior & $40 m^2$ & 1         & 2       \\  
        2  & Veranda 1     & Exterior & $25 m^2$ & 0         & 0       \\    
        3  & Interior Hall & Interior & $5 m^2$  & 0         & 0       \\    
        4  & W.C.          & Interior & $4 m^2$  & 0         & 1       \\    
        5  & Bathroom      & Interior & $6 m^2$  & 0         & 1       \\    
        6  & Bedroom 1     & Interior & $15 m^2$ & 0         & 0       \\    
        7  & Bedroom 2     & Interior & $12 m^2$ & 0         & 0       \\  
        8  & Veranda 2     & Exterior & $15 m^2$ & 0         & 0       \\  
        9  & Kitchen       & Interior & $14 m^2$ & 0         & 1       \\
        10 & Balcony       & Exterior & $4 m^2$  & 0         & 0       \\
        %\hline
    \end{tabular}
    \caption{Space-Units included in the DS for the experiment.}
    \label{tab__space_units}
\end{table}

\subsection{Algorithm Parameters}\label{sec:protocol_parameters}

For this experiment, we partition each archive into 4,096 cells ($64\times64$). BCs, feasible and infeasible fitness are as in Section \ref{sec:usecase}. The selection window is of size $9\times9$. The algorithm is initialized with 100 individuals, and FI-MAP-Elites runs until 1\% of the feasible archive is covered. All runs across all experiments start from the same initial population produced above. Between user selections, UC-ME operates within the selection window by iteratively selecting 10,000 parents (alternating between feasible and infeasible archives) before the next batch of design alternatives are shown to the user.

\subsection{Artificial Users}\label{sec:protocol_users}

In order to test the performance of our interactive QD approach, we need a way to control the user's selection criteria. We thus developed several artificial users: each artificial user ($U_i$) following a heuristic ($h_i$) as its user selection criterion (USC) when selecting among the presented alternatives. All agents select the individual with the highest $h_i$ (maximization problem) and all $h_i\in[0,1]$. Eight artificial agents ($U_1$ to $U_{8}$) have a consistent USC throughout evolution), while four artificial agents ($U_{9}$ to $U_{12}$) change their USC after 5 selections ($s$) in order to test how the algorithm adapts to a shifting user taste. This method for evaluating algorithms dependent on personal taste via artificial users (including a changing USC) is based on \cite{liapis2013rie}. We have chosen USCs that are captured in the two behavioral characterizations of our case study (BC1 and BC2), as the proposed method operates best when the user's taste is not orthogonal to the dimensions of QD explored. The heuristics of the different users are listed below:

\begin{itemize}
    \item $U_1$ maximizes BC1: $h_1=\bar{C}_s$.
    \item $U_2$ maximizes BC2: $h_2=\bar{O}_\theta$.
    \item $U_3$ maximizes both BC1 and BC2: $h_3=\frac{1}{2}(\bar{C}_s+\bar{O}_\theta)$.
    \item $U_4$ maximizes BC1 or BC2: $h_4=max(\bar{C}_s,\bar{O}_\theta)$.
    \item $U_5$ minimizes BC1: $h_5=1-\bar{C}_s$.
    \item $U_6$ minimizes BC2: $h_6=1-\bar{O}_\theta$.
    \item $U_75$ minimizes both BC1 and BC2: $h_7=1-\frac{1}{2}(\bar{C}_s+\bar{O}_\theta)$.
    \item $U_8$ minimizes BC1 or BC2: $h_8=1-max(\bar{C}_s,\bar{O}_\theta)$.
    \item $U_9$ maximizes BC1, then minimizes it: \\ 
    $h_{9}=\begin{cases}\bar{C}_s \quad \text{if $s\leq5$}\\1-\bar{C}_s \quad \text{if $s>5$}\end{cases}$
    \item $U_{10}$ maximizes BC2, then minimizes it: \\ 
    $h_{10}=\begin{cases}\bar{O}_\theta \quad \text{if $s\leq5$}\\1-\bar{O}_\theta \quad \text{if $s>5$}\end{cases}$
    \item $U_{11}$ maximizes BC1, then maximizes BC2: \\ 
    $h_{11}=\begin{cases}\bar{C}_s \quad \text{if $s\leq5$}\\
    \bar{O}_\theta \quad \text{if $s>5$}\end{cases}$
    \item $U_{12}$ maximizes BC2, then maximizes BC1: \\ 
    $h_{12}=\begin{cases}\bar{O}_\theta \quad \text{if $s\leq5$}\\ \bar{C}_s \quad\text{if $s>5$}\end{cases}$
\end{itemize}

\subsection{Performance Metrics}\label{sec:protocol_metrics}

We follow the literature for assessing the performance of MAP-Elites algorithms through observing the Coverage (percentage of occupied cells), Maximum Fitness (highest fitness among elites) and QD-Score (total fitness of all elites) of the feasible archive \cite{mouret2015illuminatingSS}. Specifically, we examine the Area Under Curve (AUC) of these metrics from the start of evolution until the end, thus measuring performance during the entirety of the run---not just the final state.
%
%Coverage, which is calculated as the percentage of occupied cells, summarizes the algorithm's global diversity capabilities. QD-Score, which is calculated as the fitness sum of all occupied cells is an attempt of capturing both Quality and Diversity of the generated results in one measure. Max Fitness [...].
%

Since we want to achieve a user-controllable exploration of the problem space, we use the user selection criterion (USC) in several variations to assess how the algorithm caters for a user's tastes. The following metrics capture whether the elites match the user's selection criteria: the maximum and average value of the USC of elites in the archive (Max USC, Mean USC). In addition, we want to assess whether good elites according to the USC are also of good quality: to measure this, we multiply the USC score of feasible elites with their fitness to derive W-USC. Inspired by the QD score, we assess both the average W-USC in the archive (Mean W-USC), and the sum of W-USC scores of feasible elites (Sum W-USC). These four metrics are evaluated in every time-step of the algorithm, and we calculate the AUC score for each of them, similarly to above.

Given the many DAS methods in Section \ref{sec:interactive_design_alternatives}, it would be important to evaluate how useful the design alternatives are to the user.
We thus introduce the following metrics: \emph{Local Diversity} as the average distance between concurrently presented alternatives in the feature space, \emph{Local Mean Fitness} as their average fitness, \emph{Local Mean USC} as their average USC value. As above, the AUC of these metrics is used for comparisons between DAS methods.

Finally, the DAS method can impact the efficiency with which the selection window reaches areas of the search space with high USC scores, versus randomly moving around. We evaluate \emph{USC Efficiency} via Eq. \eqref{eq:usc_efficiency}; it essentially captures the degree to which consecutive designer selections lead to a higher USC than was previously acquired. Unlike all other metrics, efficiency captures the entirety of the evolutionary run and thus does not need to be captured over time (as AUC).
\begin{equation}\label{eq:usc_efficiency}
\text{USC Efficiency} = \frac{\sum_{s=2}^{N_s} ( U_s - U_{s-1} ) }{\sum_{s=2}^{N_s}|U_s - U_{s-1}|}
\end{equation}
\noindent where $N_s$ is the total number of user selections in the experiment, $U_s$ is the USC value of the selected individual at selection $s$.

\section{Results}\label{sec:results}

In order to summarize our findings from experiments with the use case of Section \ref{sec:usecase}, we first compare results of UC-ME with different DAS methods (Section \ref{sec:results_das}), then we choose the most promising DAS method among them to compare against MAP-Elites without user control (Section \ref{sec:results_baseline}). Results are from 10 independent runs, and significance is established via Student's $t$-test with $p<0.05$.

\subsection{Comparisons between DAS methods}\label{sec:results_das}

%%%%%%%%%%%%%%%%%%%%%%%%%%%%%%%%%%%%%%%%%%%%%%%%%%%%%
% OSM VS OSMs table
%%%%%%%%%%%%%%%%%%%%%%%%%%%%%%%%%%%%%%%%%%%%%%%%%%%%%
\begin{table}[h]
\small
    \centering
    \begin{tabular}{l|c|c|c|c|c|c}
        %\hline
        Parameter & $A_R$ & $A_Q$ & $A_E$ & $A_S$ & $A_C$ & $A_M$ \\
        \hline
        Local Diversity &	0 &	0 &	48 &	0 &	58 &	13 \\
        Local Mean Fitness &	17 &	2 &	0 &	2 &	0 &	31 \\
        Local Mean USC &	0 &	2 &	2 &	1 &	7 &	0 \\
        \hline
        USC Efficiency     &	0 &	10 &	10 &	10 &	7 &	6 \\
        
        %\hline
    \end{tabular}
    \caption{Comparison between all DAS Methods for local QD metrics. Values show how many times this DAS method was significantly better ($p<0.05$) than another DAS method in the same experiment. Results are collected after 10 selections.}
    \label{tab:osm_vs_osm}
\end{table}

Table \ref{tab:osm_vs_osm} summarizes a comparison between different DAS methods. Treating each artificial user as a separate experiment, we evaluate in how many pairwise comparisons the DAS method had significantly better metrics than another method, after 10 selections. With 6 DAS methods (i.e. 5 comparison per method) and 12 artificial users, the maximum number in each cell is 60. The Bonferroni correction \cite{dunn1961comparisons} is applied for multiple comparisons. Note that we also tested for all other metrics (pertaining to QD and USC), but there was almost no difference between the DAS methods.

Table \ref{tab:osm_vs_osm} indicates that the Edges and Corners DAS methods have a clear advantage in local diversity. This is expected, as both methods prioritize individuals that are as far away from the selection window's center as possible. Intuitively the Corners method is slightly better at local diversity as its first choices have the absolute maximum distance of all candidates in the window (see Fig.~\ref{fig:interactive_design_alternatives}).

In terms of local mean fitness, we can see that the most successful methods in that regard are Medoids, followed by Random DAS. Intuitively, both of these methods are less likely to move the window quickly away from its current position (as indicated by their poor efficiency) and thus more QD iterations are targeting the same cells which is likely to lead to higher fitness within the window. Moreover, individuals at the edges of the feasible space tend to have a lower fitness (see Fig.~\ref{fig:heatmaps}) and thus Edges or Corners DAS methods are likely to sample at least some less fit individuals.

We also note that in terms of local mean USC and USC efficiency there are no clear winners, with $A_E$, $A_S$ and $A_M$ being slightly more efficient than other methods. Even before 10 selections, $A_C$ tends to reach the edge of the feasible space with the best USC (see Fig.~\ref{fig:heatmaps}) and after that the window moves erratically---and inefficiently.

Based on the findings of Table \ref{tab:osm_vs_osm}, the Corners DAS method has the best performance due to a higher diversity of shown individuals, while still being fairly efficient. The Medoids method shows fitter individuals to the user than other DAS methods, while still being somewhat efficient at adapting to the USC. We thus test these DAS methods against a baseline MAP-Elites in the next Section.

\subsection{Comparisons with MAP-Elites}\label{sec:results_baseline}

\begin{table}[h]
\small
\centering
\begin{tabular}{l|c|c||c|c}
%\hline
Parameter &	MAP-Elites & $A_C$ & MAP-Elites & $A_M$ \\
\hline
Coverage &	12 &	0 &	12 &	0 \\
Max Fitness &	0 &	0 &	0 &	0 \\
QD Score &	12 &	0 &	12 &	0 \\
\hline
Max USC & 1 &	9 &	1 &	4 \\
Mean USC &	0 &	12 &	0 &	10 \\
Mean W-USC & 0 &	11 &	0 &	11 \\
Sum W-USC &	11 &	0 &	12 &	0 \\
%\hline
\end{tabular}
\caption{Comparison between baseline MAP-Elites and two of the best performing DAS methods, showing the number of experiments where one method or the other had significantly higher scores  ($p<0.05$) in each metric. Results are collected after 10 selections.}
\label{tab:osm_vs_baseline}
\end{table}

\begin{figure*}
    \includegraphics[width=.99\textwidth]
        {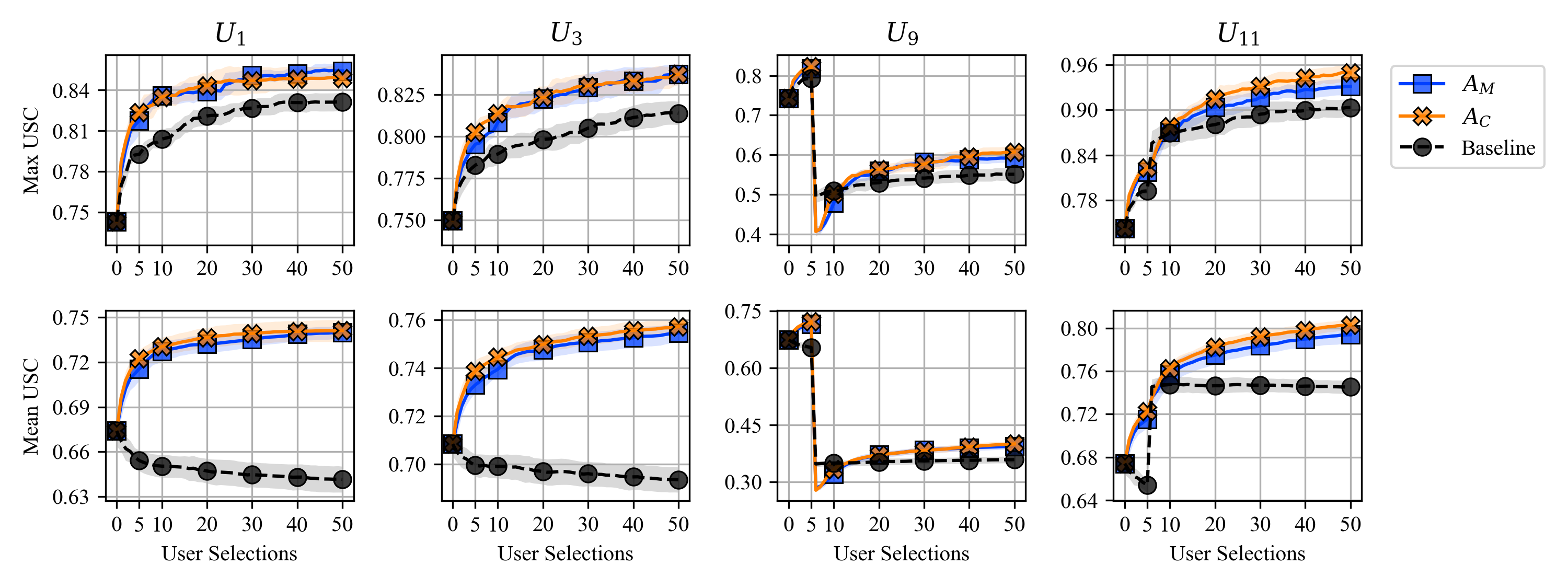}
    \caption{Charts display the value of Max and Mean USC of four different artificial users ($U_1$, $U_4$, $U_9$ and $U_{11}$), comparing the Quadrants ($A_C$) and Medoids ($A_M$) DAS methods with the baseline (MAP-Elites without user control). Values are averaged across 10 different runs and shaded regions capture the 95\% confidence interval.}
    \label{fig:usc_charts}
\end{figure*}

Based on the comparisons between DAS methods in Section \ref{sec:results_das}, we focus on comparing the Corners and Medoids methods against a baseline MAP-Elites which does not consider the user's taste and performs unguided exploration of the search space. The baseline implements FI-MAP-Elites \cite{sfikas2022aGP} and randomly selects random individuals to mutate, alternating between the two archives.

Table \ref{tab:osm_vs_baseline} shows the pairwise comparison between the unguided MAP-Elites baseline and each UC-ME variant. Results show the number of experiments in which the baseline or the UC-ME variant had superior performance in terms of the chosen metric (out of a total of 12 experiments), after 10 user selections. It is evident that unguided MAP-Elites has better coverage of the problem space and thus a higher QD score, across all experiments. This is not surprising, as UC-ME drives search towards specific parts of the problem space (and regions of the feature map), while MAP-Elites covers as much of the feature map as possible. We also note that there are no differences in terms of maximum fitness. This is somewhat surprising, since different parts of the feature map (targeted by different users) may not have equally good fitnesses. It seems that finding a highly fit individual is not challenging in this use case.

As expected, the unguided exploration of the baseline MAP-Elites performs worse than both UC-ME versions for maximum and mean USC score of all elites in the archive. The $A_M$ method is less efficient at reaching very high USC scores, compared to $A_C$; this is not surprising since the latter moves the selection window toward regions of the problem space with high USC faster. Mainly due to a higher mean USC, it is not surprising that the mean W-USC is higher for UC-ME variants compared to the baseline. The higher coverage of the baseline, however, leads to higher values in the sum of W-USC scores among all elites, similarly to the QD Score.

Figure \ref{fig:usc_charts} shows a comparison between the progression of mean USC and max USC for four indicative artificial users: two consistent ($U_1$, $U_3$) and two that change criteria after 5 selections ($U_9$, $U_{11}$). Note that throughout Section \ref{sec:results} we chose to focus on a threshold of 10 selections because this is a realistic number of times that the user may interact with a tool such as this one before user fatigue kicks in. However, for completeness we let experiments run until 50 user selections were completed and display the full runs in Fig.~\ref{fig:usc_charts}. 

It is evident from Fig.~\ref{fig:usc_charts} that while the unguided MAP-Elites can accidentally find regions of the problem space with a high USC (i.e. max USC keeps increasing), this is not a guarantee for the broader population (mean USC may increase or decrease) depending on which parts of the space are more easily reachable. As expected, UC-ME variants consistently improve both USC measures as the archive is driven by local QD towards specific parts of the space. 

Figure \ref{fig:usc_charts} also shows how the algorithms handle abrupt changes in user criteria after the 5th selection ($U_9$, $U_{11}$). $U_9$ has a more abrupt change as it suddenly targets the opposite of its previous criterion, causing a drop in USC. It takes several user interactions to move the window towards more appropriate regions of the problem space, but after 5 selections from the criterion change, UC-ME approaches the mean and max USC of the baseline which has been evolving for both high $\bar{C}_s$ and low $\bar{C}_s$ (both captured in $h_9$). Given enough time, both methods surpass the baseline (e.g. after 25 selections). $U_{11}$ is not as ``aggresssive'' in changing its mind; indeed, even unguided MAP-Elites can find individuals with high $\bar{O}_\theta$, which is the USC from 6th selection onward. Since the selection window does not have to retrace its steps, as with $U_9$, the UC-ME methods can find comparable or slightly better individuals to the baseline after 5 more selections, and much better individuals given enough time.

\begin{figure}
\centering
\includegraphics[width=0.8\textwidth]{    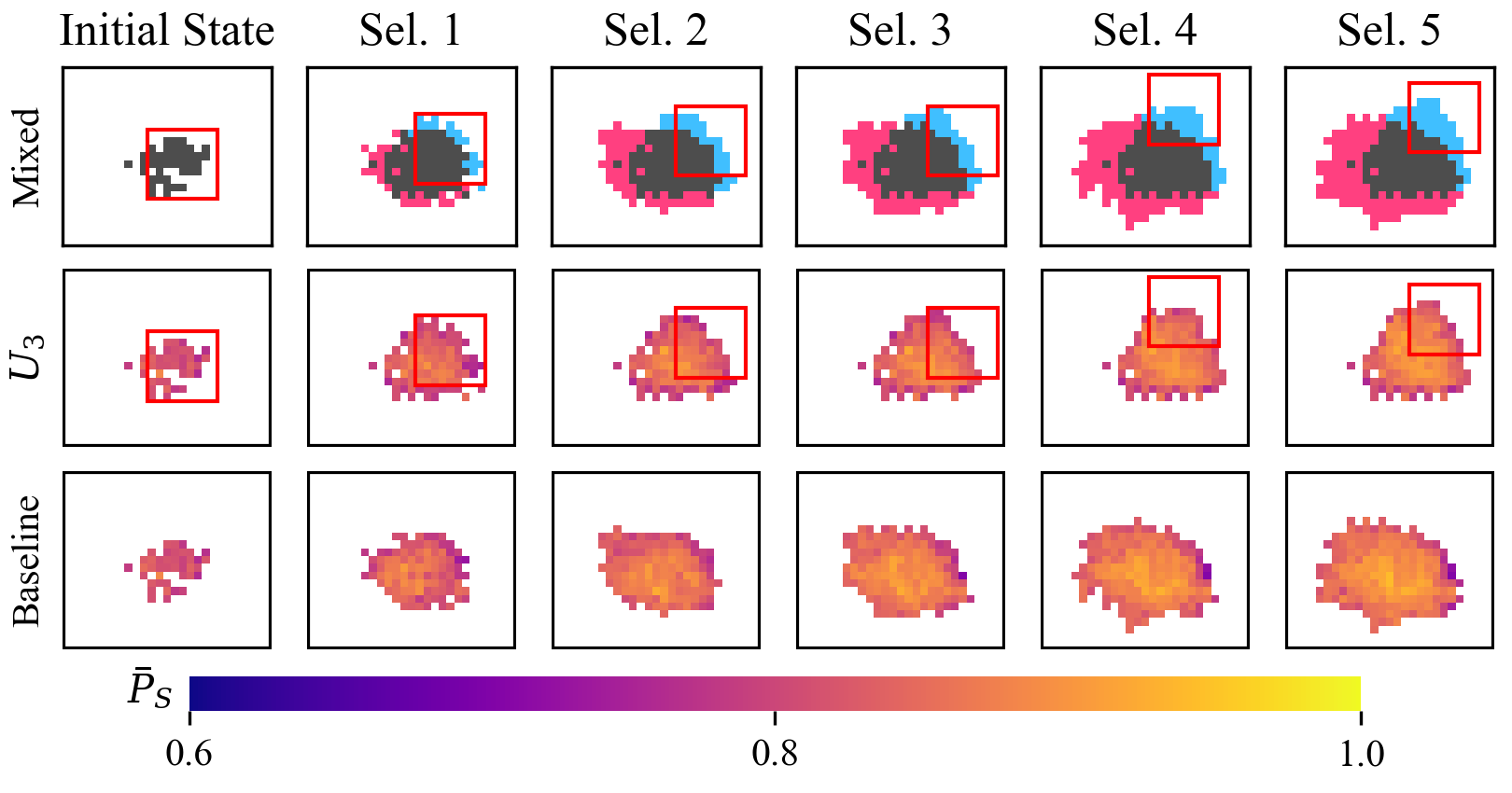}
%  }
  \caption{Behavioral space exploration for the baseline MAP-Elites (bottom row) and UC-ME with Corners DAS guided by $U_3$ (middle row), for the first 5 selections. Their shared color scale (shown at the bottom) represents fitness $\bar{P}_s \in [0.6, 1]$. The top row shows coverage differences: red cells are discovered only by the baseline, blue cells are discovered only by UC-ME and gray cells are common. In these figures the $x$ axis is $\bar{C}_s\in[0.44, 0.86]$ and the $y$ axis is $\bar{O}_\theta\in[0.61, 0.97]$.}
  \label{fig:heatmaps}
\end{figure}

The progress of UC-ME can also be visualized through the feature map itself. Figure \ref{fig:heatmaps} shows how coverage changes after each user selection (or the same evaluation threshold for MAP-Elites). In addition, the figures show in red the selection window of UC-ME as it moves towards higher USC scores (in this case that of $U_3$). We focus on the $A_C$ method, as the most efficient. The top row of images in Fig.~\ref{fig:heatmaps} illustrates the differences between UC-ME and MAP-Elites exploration patterns: in gray we see the common cells discovered by both methods, in magenta we see the cells discovered only by MAP-Elites and in blue we see the cells discovered only by UC-ME. We see that cells at higher USC values exclusively belong to UC-ME. The higher coverage of MAP-Elites is due to most cells occupying lower $\bar{C}_s$ and $\bar{O}_\theta$ values, which are undesirable for $U_3$. Figure \ref{fig:heatmaps} also shows how the selection window moves first towards a higher $\bar{C}_s$; once it reaches the edge of the feasible space and can not find individuals with higher scores in that direction, it moves towards higher $\bar{O}_\theta$ scores. We also see that within the first 3 selections, UC-ME with $A_C$ has found the edges of the feasible space with the highest USC scores and starts moving around fairly haphazardly in that vicinity, leading to more selections and improved quality of individuals in that specific region of the problem space.

\newcommand{\sfw}{0.26\textwidth}
\begin{figure}
	\centering
	\subfloat[Selection 1]{
		\label{fig:b}
		\includegraphics[width=\sfw]{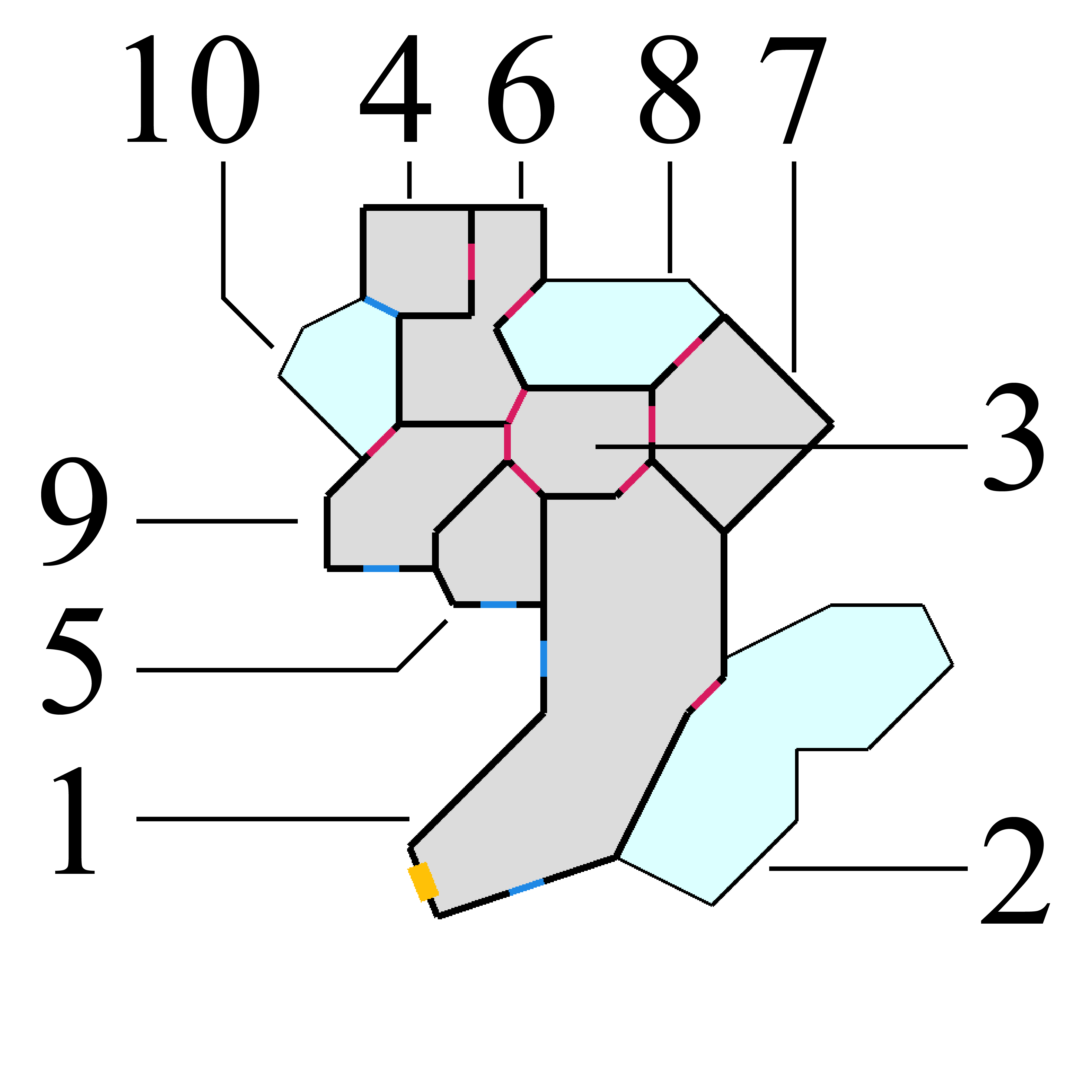}
        }
	\qquad
	\subfloat[Selection 10]{
		\label{fig:c}
		\includegraphics[width=\sfw]{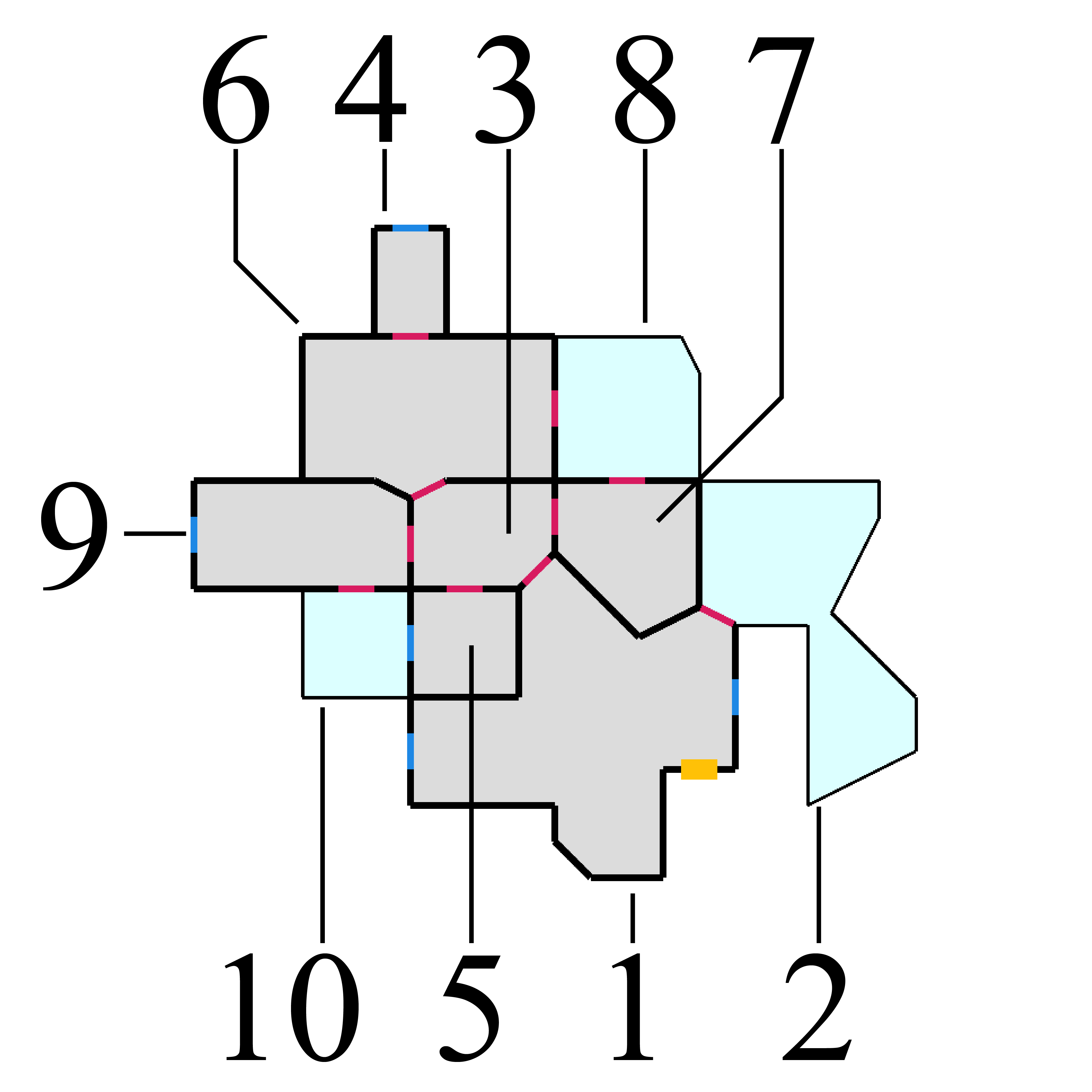}
        }
    \qquad
	\subfloat[Selection 50]{
		\label{fig:d}
		\includegraphics[width=\sfw]{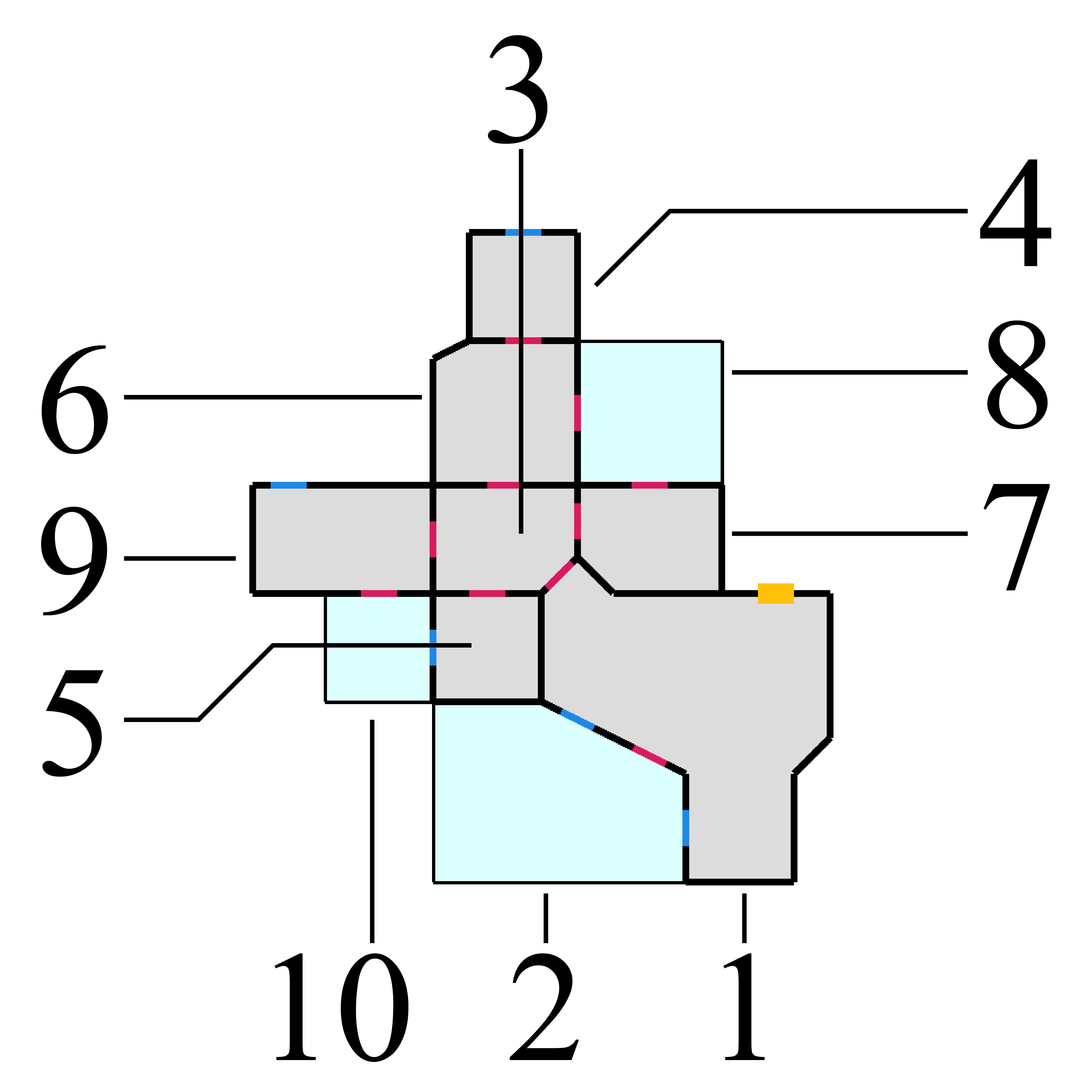}
        }
	\caption{Indicative individuals selected by $U_3$, using the $A_C$ DAS method, after 1, 10 and 50 interactions with UC-ME.}
	\label{fig:examples}
\end{figure}

Finally, we show what $U_3$ selected in an indicative run of UC-ME with $A_C$ in Fig.~\ref{fig:examples}. We observe that initial individuals do not have a good USC score as the shown selection has many acute angles and complex shapes in the rooms. After 10 selections, the user has found an individual with mostly compact and square rooms. After 50 selections, the results are not much different than with 10 selections; thus 10 selections are usually enough for this problem and would not be overly fatiguing to the user.

\section{Discussion}\label{sec:discussion}

This paper considers that the space partitioning of MAP-Elites can benefit interactive evolution: the (moving) selection window is a simple and intuitive way to localize QD search, treating the window (and the BCs) as an implicit function approximator \cite{takagi2001interactive}. Our experiments have showcased that while the unguided MAP-Elites explores more of the problem space, UC-ME finds individuals more pertinent to the user's taste and covers fairly large regions of the feasible space that have desirable characteristics (see Fig.~\ref{fig:heatmaps}). Interestingly, experiments with artificial users did not reveal major differences between DAS methods; however, we expect that different shown alternatives may impact human users more, making their selection harder and increasing fatigue.

Therefore, an intuitive next step is to deploy UC-ME to human designers, with experience in architectural layouts, and evaluate both the performance metrics listed in Section \ref{sec:protocol_metrics} and the users' experience through a post-study usability questionnaire \cite{lewis1990integrated}. Unlike artificial users, the complex problem of architectural layouts may be difficult to process for human users, while also choosing between more or less similar options. Therefore, future work should explore how UC-ME performs in unconstrained (or less constrained) problems, including problems which require less cognitive effort on the part of the human user to process the results visually. An intuitive unconstrained case study would be image generation---to which MAP-Elites has already been applied \cite{fontaine2021differentiable}. Other directions for future work could explore constrained domains such as games where QD search can be beneficial to give game designers diverse but good options \cite{gravina2019pcgqd}; possible cognitive-light tasks in the games domain could be sprite generation \cite{liapis2016arcade} or narrative structures \cite{alvarez2022tropetwist}.

Beyond testing in other domains and with human users, there is a plethora of future directions for improving the algorithmic properties of UC-ME. Specifically, implementing a variant that operates within a higher-dimensional feature space (more than 2 dimensions) would be a worthwhile step, as it would offer the user more control over the search direction. Such an approach could leverage CVT-MAP-Elites \cite{vassiliades2018cvt}, which can theoretically operate on more dimensions, and convert the selection window into a point cloud (of nearest CVT neighbors) moving within that space. A more ambitious avenue of future work would be the re-introduction of user models \cite{liapis2013designermodeling} to UC-ME, as a complementary way of (aesthetic) function approximation \cite{pei2018researchPS} to the local QD. Moreover, a constantly updated user model \cite{hagg2019modeling} could potentially be used to adapt the behavior characterizations themselves to better align to the user's taste, enabling a simpler UC-ME pipeline that moves the selection window around. This ambitious direction could have several applications beyond UC-ME, and could leverage existing methods of automatic and incremental behavior characterization \cite{cully2019aurora,hagg2018prototype,hagg2021expressivity}.

\section{Conclusion}\label{sec:conclusion}

In this paper, we propose a way of controlling the direction of exploration and the computational budget of Quality Diversity search with little cognitive load. The user-controlled MAP-Elites algorithm (UC-ME) is the first instance of the interactive evolution paradigm (in its narrow sense) applied to MAP-Elites, unlike past work \cite{alvarez2020interactive}. The UC-ME algorithm operates by focusing selection of parents in a smaller window of the feature map, which allows it to operate in both unconstrained and constrained problems via two archives \cite{sfikas2022aGP}. In our experiments, on a complex and heavily constrained problem, we observe that UC-ME can focus on interesting parts of the problem space according to the user's selection criterion, at the cost of lower coverage and fewer elites in total. The proposed method shows potential, but important next steps include testing its efficiency and impact on user fatigue with human users, as well as expanding the work in more domains, with more BC dimensions, and alongside constantly updated models of the user's taste \cite{liapis2013designermodeling}. 

\section{Acknowledgments}

This work has been supported by the European Union’s Horizon 2020 research and innovation programme from the PrismArch (Grant Agreement No. 952002) and AI4media  (Grant Agreement No. 951911) projects. 

\bibliographystyle{unsrt}
\bibliography{ucme_references}

\end{document}